\documentclass[]{fairmeta}

\usepackage{multirow}
\usepackage{graphicx}
\usepackage{array}
\usepackage{hyperref}
\usepackage{amsmath,amssymb,amsfonts}
\usepackage{cleveref}
\usepackage{subcaption}
\usepackage[numbers]{natbib}
\usepackage{booktabs}
\usepackage{enumitem}
\usepackage{fontawesome5}

\newcommand{\commentout}[1]{}
\renewcommand{\paragraph}[1]{\noindent\textbf{#1}\hspace*{1em}}
\setlist[itemize]{leftmargin=15pt}

\title{ReflectRL: Learning from Golden Negative Trajectories via Reflective-to-Direct Reasoning}

\author[1,2,6,9]{Jinhe Bi}
\author[3]{Chennan Zhou}
\author[2]{Zengjie Jin}
\author[2,9]{Aniri}
\author[8]{Shuo Lu}
\author[4]{Wenke Huang}
\author[7]{Hu Cao}
\author[6]{Xun Xiao}
\author[5]{Zhihong Zhu}
\author[2]{Volker Tresp}
\author[1]{Fei Shen}
\author[2,9,\mbox{\scriptsize\faEnvelope}]{Yunpu Ma}
\author[1]{Tat-Seng Chua}

\affiliation[1]{National University of Singapore}
\affiliation[2]{Ludwig Maximilian University of Munich}
\affiliation[3]{Technical University of Munich}
\affiliation[4]{Nanyang Technological University}
\affiliation[5]{Peking University}
\affiliation[6]{Huawei Heisenberg Research Center}
\affiliation[7]{Southeast University}
\affiliation[8]{Institute of Automation, Chinese Academy of Sciences}
\affiliation[9]{Munich Center for Machine Learning}

\vspace{0.5em}
\contribution[]{\href{https://github.com/bibisbar/ReflectRL}{\faGithub~ GitHub} \quad \href{https://huggingface.co/datasets/Jinhe/OpenR1-GNT-69k}{\faDatabase~ Dataset} \quad \href{https://huggingface.co/Jinhe/ReflectRL-Qwen2.5-Math-7B-GRPO-ReflectRL}{\faCube~ Model}}

\abstract{
On-policy training has emerged as a powerful post-training paradigm for improving the reasoning capabilities of large language models, and is often enhanced by golden trajectories from stronger expert models. However, when the expert fails on harder problems, existing trajectory-guided methods lose their main source of supervision, and these failed trajectories are typically discarded as negative samples. We argue that such failures, which we call Golden Negative Trajectories, can still provide valuable reasoning signals when treated not as demonstrations to imitate, but as flawed trajectories to reflect upon.
We identify a Reflection Advantage: for hard problems, reflecting on a flawed trajectory can be easier and more effective than solving the problem directly from scratch. Motivated by this, we propose ReflectRL, a lightweight plug-and-play framework that learns from Golden Negative Trajectories during on-policy training. ReflectRL first uses these trajectories to elicit Reflective Reasoning, then applies Reflective-to-Direct Policy Transition to transfer the acquired reasoning behavior back to Direct Reasoning. Experiments across 9 benchmarks, 4 LLM backbones, and 4 on-policy training methods show that ReflectRL consistently improves reasoning performance with minimal overhead.
}

\date{August 2026}
\correspondence{\email{bijinhe@outlook.com}}

\begin{document}
\thispagestyle{firstheader}
\maketitle

\section{Introduction}

On-policy training has become a central recipe for improving the reasoning capabilities of large language models \cite{DeepSeekAI2025DeepSeekR1IR,yu2026dapo,Bi2025CoTKineticsAT,huang2025loongsynthesizelongchainofthoughts,mid,ma-etal-2026-self}. In this paradigm, a model samples reasoning rollouts on training problems and updates its policy using feedback from verifiers, reward models, or distributional objectives. Recent methods such as Reinforcement Learning with Verifiable Rewards (RLVR) and On-Policy Distillation (OPD) show that this training paradigm can substantially improve reasoning performance when the sampled rollouts provide informative learning signals \citep{lu2025onpolicydistillation,DeepSeekAI2025DeepSeekR1IR}. This dependence on rollout quality becomes more critical on hard problems, where direct sampling often produces sparse positive rewards and weak policy updates \citep{bi2026echorl}.

A common strategy for improving on-policy training is to incorporate golden reasoning trajectories from stronger expert models, such as DeepSeek-R1 \citep{DeepSeekAI2025DeepSeekR1IR}. In RLVR, correct experts' golden trajectories can guide rollout generation toward more promising reasoning paths \citep{NEURIPS2025_a9d5c33e,liu2026uft}. In OPD, golden trajectories from expert models can serve as privileged information for shaping the learning signal \citep{zhao2026selfdistilledreasoneronpolicyselfdistillation}. These trajectories are most useful when the expert model solves the problem and provides a correct reasoning trajectory. When the expert model fails on harder problems, existing trajectory-guided methods lose their main source of positive supervision. The failed expert trajectories are then typically filtered out as negative samples, leaving a large amount of structured reasoning from strong models unused. This deficiency in the prior art brings up an open question:

\textit{Is it possible to leverage these discarded but still informative negative trajectories to continuously improve the reasoning capability?}

\begin{figure*}[t!]
\centering
\includegraphics[width=\textwidth]{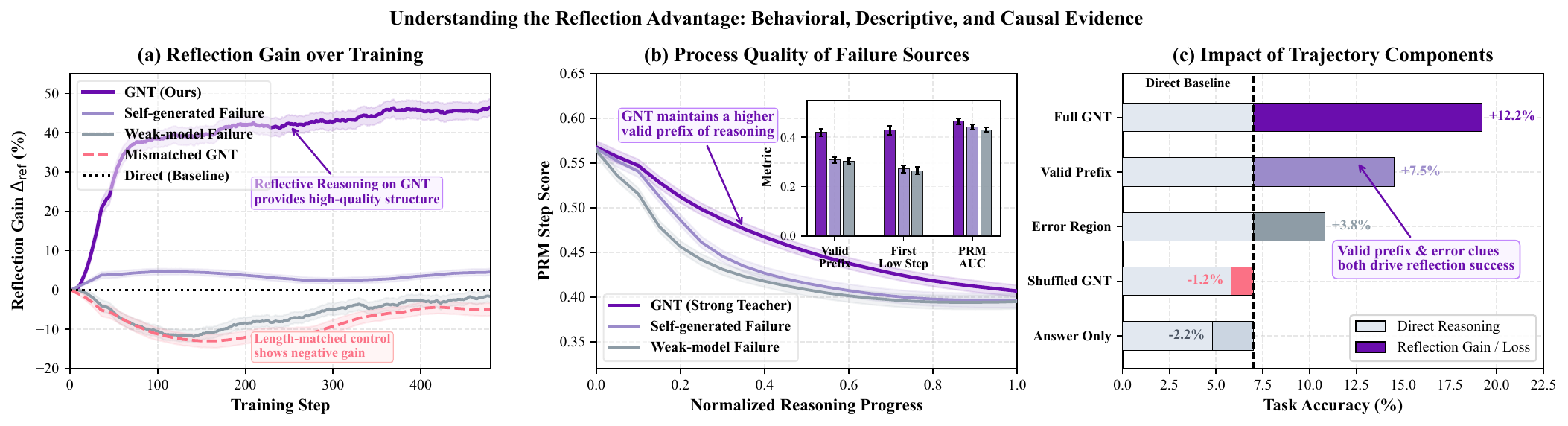}
\caption{
Understanding the \textbf{Reflection Advantage} through behavioral, descriptive, and intervention evidence on Qwen2.5-Math-7B.
(a) \textbf{Behavioral Evidence}: Reflection Gain $\Delta_{\text{ref}} = \mathbb{E}_{o \sim \pi}[r(q, o) \mid q, o^-] - \mathbb{E}_{o \sim \pi}[r(q, o) \mid q]$ over training steps. Reflecting on a GNT yields the largest gain, which steadily grows during training, whereas self-failures offer limited benefits, and weak-model failures or mismatched GNT (incase length bias) leads to negative gain.
(b) \textbf{Descriptive Evidence}: Qwen2.5-Math-PRM-7B score profiles of failure sources and process metrics (inset). GNT maintains a higher valid prefix of reasoning compared to other failure sources before dropping at localized error points.
(c) \textbf{Intervention Evidence}: Ablation of GNT trajectory components. Both the valid prefix and the error region causally drive the reflection gain, whereas removing these components (Shuffled GNT, Answer Only) reduces accuracy below the direct baseline.
}
\label{fig:reflection_advantage}
\end{figure*}

\begin{figure*}[t]
\centering
\includegraphics[
    width=\linewidth
]{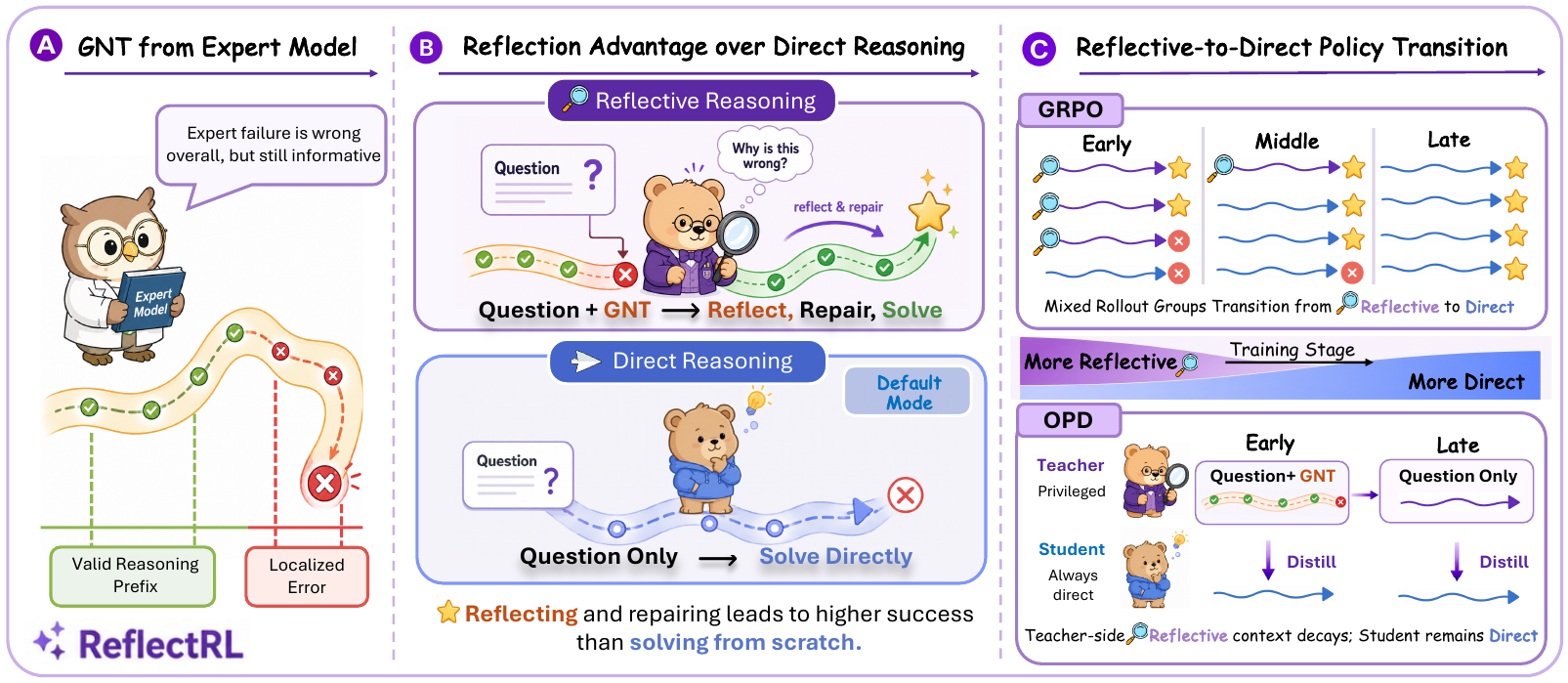}
\caption{
Overview of \textbf{ReflectRL}. A failed expert trajectory is used as a \textbf{Golden Negative Trajectory} to elicit \textbf{Reflective Reasoning}. ReflectRL combines Reflective Reasoning with \textbf{Direct Reasoning} during on-policy training and applies \textbf{Reflective-to-Direct Policy Transition} to transfer the acquired reasoning behavior back to the original problem setting.
}
\label{fig:teaser}
\end{figure*}

\textbf{The Present Work: }
ReflectRL addresses this question by exploiting these valuable negative trajectories, which are typically discarded under existing training paradigms \citep{liu2026uft,NEURIPS2025_a9d5c33e}.
However, incorporating negative samples into the current on-policy training paradigm is fundamentally challenging. Specifically, trajectory-guided RLVR variants rely on correct expert trajectories to yield positive rewards, establishing a relative advantage over self-generated negative rollouts. Because negative trajectories from expert models fail to produce positive rewards, these methods cannot exploit them. Consequently, standard advantage estimation cannot distinguish the structured, high-quality expert failures from the model's own low-quality self-failures, preventing these negative trajectories from being leveraged to drive policy updates. Alternatively, when external trajectories are used as privileged information in OPD paradigms, they are typically designed as positive demonstrations to be imitated, which directly contradicts the flawed nature of negative trajectories.

To address this gap, we first conduct an empirical analysis to locate where usable learning signals reside in expert-generated negative trajectories and investigate how the target model can leverage them. As shown in Figure~\ref{fig:reflection_advantage}(a), we identify a distinct behavioral pattern: compared to direct reasoning on hard questions without any external context—the default mode in on-policy training, the model achieves a substantial reasoning accuracy improvement (which we term \emph{Reflection Gain}) when it is hinted with a negative trajectory and forced to reflect on the errors.
We define this pattern as the \emph{Reflection Advantage}. Importantly, we find that negative trajectories generated by expert models (which we refer to as \emph{Golden Negative Trajectories}-GNTs) yield a far more pronounced Reflection Advantage than failure trajectories from the target model itself or a weaker model as shown in Figure~\ref{fig:reflection_advantage}(a). To understand why GNTs are uniquely beneficial, we utilize a Process Reward Model (PRM) \citep{Zhang2025TheLO} to profile the step-level quality of different negative trajectory sources. As shown in Figure~\ref{fig:reflection_advantage}(b), GNTs maintain higher process rewards and longer valid reasoning prefixes before encountering localized errors. To isolate this signal, we perform causal interventions on GNT components (Figure~\ref{fig:reflection_advantage}(c)), demonstrating that both the valid prefix and the localized error region jointly drive the Reflection Advantage. This suggests that the model not only benefits from continuing reasoning along a high-quality prefix but also learns to actively reflect on and recover from specific errors.

Leveraging this Reflection Advantage, as shown in Figure~\ref{fig:teaser}, we propose \textbf{ReflectRL}, a lightweight framework for learning from GNTs during on-policy training. ReflectRL keeps the base training objective unchanged and only modifies how training rollouts are constructed. For RLVR, ReflectRL samples both Reflective Reasoning rollouts and Direct Reasoning rollouts, evaluates them with the same verifier, and applies the standard policy update. For OPD, ReflectRL incorporates GNTs as privileged context on the teacher-side, while keeping the student aligned with Direct Reasoning. This unified design allows ReflectRL to be easily integrated into various on-policy training paradigms.

Reflective Reasoning enables the model to learn from GNTs during training, while Direct Reasoning remains the desired behavior at inference time. ReflectRL connects these two reasoning modes through \textbf{Reflective-to-Direct Policy Transition}. During training, the rollout distribution is gradually shifted from Reflective Reasoning toward Direct Reasoning. This transition encourages the model to preserve the reasoning behavior learned from GNTs while reducing its dependence on expert models.


We evaluate ReflectRL in representative on-policy training settings covering 9 benchmarks, 4 LLM backbones, and 4 training methods, including both RLVR and OPD. The results show that ReflectRL consistently improves reasoning performance with minimal overhead. Further analysis supports the Reflection Advantage and shows that Reflective-to-Direct Policy Transition helps transfer the benefit of Golden Negative Trajectories back to Direct Reasoning.

Our contributions are summarized as follows:
\begin{itemize}
\item We introduce \textbf{Golden Negative Trajectories} as a useful source of reasoning signal for on-policy training, and release \textbf{OpenR1-GNT-69k}, a dataset consisting of 69k expert failure trajectories.
\item We identify the \textbf{Reflection Advantage}, showing that hard problems can be easier and more effective to solve through Reflective Reasoning than through Direct Reasoning.
\item We propose \textbf{ReflectRL}, a lightweight framework that integrates Golden Negative Trajectories into on-policy training through \textbf{Reflective-to-Direct Policy Transition}.
\item We demonstrate consistent improvements across representative on-policy training settings covering RLVR and OPD, with minimal additional overhead.
\end{itemize}

\section{Preliminaries}
\label{sec:prelim}
This section establishes the notation used throughout the paper and briefly reviews the GRPO and OPD objectives.
We also introduce a concise definition of Reflection Advantage, which is used to motivate our method and analysis.

\paragraph{Notation}
Let $q$ be an input query, and $\mathcal{D}$ represent a dataset of prompt-answer pairs $(q,a)$, where $a$ is the ground-truth answer.
The likelihood of generating a response rollout $o=(o_1,\dots,o_{|o|})$ given $q$ under policy $\pi_\theta$ is factorized autoregressively as:

\begin{equation}
\pi_\theta(o \mid q)
~=~
\prod_{s=1}^{|o|}
\pi_\theta\!\left(o_s \mid q, o_{<s}\right),
\label{eq:ar_policy}
\end{equation}
where $\pi_\theta$ is the trainable policy parameterized by $\theta$,
$o_{<s}\stackrel{\mathrm{def}}{=}(o_1,\dots,o_{s-1})$ denotes the history prefix, and $|o|$ denotes the response length.

\paragraph{Group Relative Policy Optimization (GRPO)}
Group Relative Policy Optimization (GRPO) is widely used as a representative method for RLVR \citep{DeepSeekAI2025DeepSeekR1IR}.
For a given prompt $(q, a)$, the algorithm samples a group of $N$ outputs $\{o^{(i)}\}_{i=1}^{N}$ from the frozen rollout policy $\pi_{\theta_{\mathrm{old}}}$.
Each rollout $o^{(i)}$ is evaluated to obtain a verifiable reward $r_i$.
Rather than employing a separate critic network, GRPO calculates the relative advantage for each rollout by normalizing the rewards within the sampled group:
\begin{equation}
\hat{A}_{i}
~=~
\frac{
r_i - \mu_r(q)
}{
\sigma_r(q)
},
\label{eq:grpo_adv}
\end{equation}
where $\mu_r(q)\stackrel{\mathrm{def}}{=}\mathrm{mean}(\{r_j\}_{j=1}^N)$ and
$\sigma_r(q)\stackrel{\mathrm{def}}{=}\mathrm{std}(\{r_j\}_{j=1}^N)$ are the mean and standard deviation of rewards within the group for query $q$,
and the advantage $\hat{A}_i$ is broadcast to each token position $s$ in rollout $o^{(i)}$.
\paragraph{On-Policy Distillation (OPD)}
On-policy distillation \citep{lu2025onpolicydistillation} transfers knowledge from a fixed teacher policy $\pi_{\mathrm{teacher}}$ to a trainable student policy $\pi_\theta$ on student-generated trajectories. Given a query $q$, the student samples a rollout $o \sim \pi_\theta(\cdot \mid q)$. At each token position $s$, OPD minimizes the reverse Kullback--Leibler (KL) divergence from the student's predictive distribution to the teacher's predictive distribution:
\begin{equation}
\begin{split}
\mathcal{L}_{\mathrm{OPD}}(\theta)
~&=~
\mathbb{E}_{q \sim \mathcal{D},\, o \sim \pi_\theta(\cdot \mid q)}
\Big[ \\
&\quad \sum_{s=1}^{|o|}
\mathbb{D}_{\mathrm{KL}}\!\left(
\pi_\theta(\cdot \mid q, o_{<s})
\,\parallel\,
\pi_{\mathrm{teacher}}(\cdot \mid q, o_{<s})
\right)
\Big].
\end{split}
\label{eq:opd_loss}
\end{equation}
By training on student-generated states rather than offline expert demonstrations, OPD alleviates the exposure bias and distribution shift issues common in offline imitation learning.

\section{ReflectRL}
\label{sec:reflectrl}
This section presents ReflectRL, a lightweight framework designed to leverage Golden Negative Trajectories (GNTs) during on-policy training, as overviewed in Figure~\ref{fig:teaser}.
Crucially, ReflectRL introduces no auxiliary loss terms, requires no changes to the outcome reward or inference-time interfaces, and preserves the original RLVR/OPD objectives.

\subsection{Direct \& Reflective Reasoning Interfaces}
\label{sec:interfaces}

To enable the policy to learn from expert failures, we define two reasoning interfaces based on distinct prompt templates. Specifically, $\text{Chat\_Temp}_D$ and $\text{Chat\_Temp}_R$ denote the prompt templates designed to activate Direct Reasoning and Reflective Reasoning, respectively.

\paragraph{Direct Reasoning}
The direct reasoning interface applies the direct prompt template $\text{Chat\_Temp}_D$ to the query $q$:
\begin{equation}
x^D(q) = \text{Chat\_Temp}_D(q).
\label{eq:direct_interface}
\end{equation}
This interface provides only the original query and prompts the policy to solve the problem directly, without access to any external trajectory. It corresponds to the default reasoning mode in standard on-policy training.

\paragraph{Reflective Reasoning}
The reflective reasoning interface applies the reflective prompt template $\text{Chat\_Temp}_R$ to the query $q$ and a pre-generated Golden Negative Trajectory $o^-$:
\begin{equation}
x^R(q, o^-) = \text{Chat\_Temp}_R(q, o^-).
\label{eq:reflective_interface}
\end{equation}
This template presents the GNT as contextual information and prompts the policy to identify its errors, repair the reasoning process, and derive a corrected solution. The detailed formats of $\text{Chat\_Temp}_D$ and $\text{Chat\_Temp}_R$ are provided in Appendix~A.


\paragraph{Reflection Advantage}
Using these interfaces, we formalize the Reflection Advantage that motivates our framework. Let $r(q, o) \in \{0, 1\}$ represent the binary outcome reward indicating the correctness of rollout $o$ for query $q$. We define the \emph{Reflection Gain} $\Delta_{\mathrm{ref}}$ of a policy $\pi_\theta$ given query $q$ and GNT $o^-$ as the expected correctness difference between reflective and direct reasoning:
\begin{equation}
\begin{split}
\Delta_{\mathrm{ref}}(\theta; q, o^-)
~=~&
\mathbb{E}_{o \sim \pi_\theta(\cdot \mid x^R(q, o^-))} \!\left[ r(q, o) \right]
\\
&-~
\mathbb{E}_{o \sim \pi_\theta(\cdot \mid x^D(q))} \!\left[ r(q, o) \right].
\end{split}
\label{eq:reflection_gain}
\end{equation}
The \emph{Reflection Advantage} describes the empirical behavior where the policy achieves a positive Reflection Gain ($\Delta_{\mathrm{ref}} > 0$). Under this condition, identifying and correcting errors in a structured expert failure is easier than generating a correct path from scratch.

\subsection{Reflective-to-Direct Policy Transition}
\label{sec:transition}
To internalize the reflective reasoning capabilities (elicited by GNTs) into the model's direct reasoning pathway for actual inference, we design a Reflective-to-Direct Policy Transition mechanism.
Without this transition, a model trained exclusively on reflective inputs would suffer from exposure bias, becoming dependent on external expert failure trajectories that are unavailable during inference.
Conversely, a smooth transition encourages the model to compile the error-correction and reasoning behaviors learned from GNTs into its own parameters, effectively transferring the reflection advantage back to the direct interface.

Specifically, at training step $t$, we define a transition kernel $g(t)$ to govern the target proportion of Reflective Reasoning. We default to cosine decay:
\begin{equation}
g(t)=p_l+\frac{p_h-p_l}{2}\left[1+\cos\!\left(\pi\tau(t)\right)\right],
\label{eq:transition_kernel}
\end{equation}
where $p_h$ and $p_l$ are the initial and terminal proportions, and $\tau(t)\in[0,1]$ is the normalized transition progress after warm-up; Appendix~C provides the exact schedule. The number of reflective rollouts is $K_t=\operatorname{round}(N g(t))$, with the remaining $N-K_t$ rollouts using Direct Reasoning. In our RLVR setting, the terminal allocation rounds to zero, yielding fully direct rollout groups.

\subsection{ReflectRL in RLVR}
\label{sec:rlvr_instantiation}
We instantiate ReflectRL in the RLVR paradigm by optimizing a joint group-relative objective.
For each training query, we generate a mixed group of $N$ rollouts $o^{(i)} \sim \pi_{\theta_{\mathrm{old}}}(\cdot \mid x_t^{(i)})$, where the prompt context $x_t^{(i)}$ is determined by the transition schedule $K_t$:
\begin{equation}
x_t^{(i)} =
\begin{cases}
\text{Chat\_Temp}_R(q, o^-), & i \le K_t, \\
\text{Chat\_Temp}_D(q), & i > K_t.
\end{cases}
\label{eq:prompt_assignment}
\end{equation}
Rather than separating the direct and reflective reasoning modes, we jointly optimize the policy by maximizing the objective $\mathcal{L}_{\mathrm{ReflectRL}}(\theta)$:
\begin{equation}
\begin{split}
\mathbb{E} \Big[ &\frac{1}{N} \sum_{i=1}^{K_t} \sum_{s=1}^{|o^{(i)}|} \min \left( w_{i,s}(\theta) \hat{A}_i, \mathrm{clip}(w_{i,s}(\theta)) \hat{A}_i \right) \\
+ &\frac{1}{N} \sum_{i=K_t+1}^{N} \sum_{s=1}^{|o^{(i)}|} \min \left( w_{i,s}(\theta) \hat{A}_i, \mathrm{clip}(w_{i,s}(\theta)) \hat{A}_i \right) \Big].
\end{split}
\label{eq:reflectrl_objective}
\end{equation}
Here, the advantage $\hat{A}_i$ is computed jointly over the mixed group according to Equation~\ref{eq:grpo_adv}, with rewards evaluated by $r(q, o^{(i)}) = \mathrm{MathVerify}(a, o^{(i)})$. The likelihood ratio $w_{i,s}(\theta)$ is defined as:
\begin{equation}
w_{i,s}(\theta) = \frac{\pi_\theta\left(o_s^{(i)} \mid x_t^{(i)}, o_{<s}^{(i)}\right)}{\pi_{\theta_{\mathrm{old}}}\left(o_s^{(i)} \mid x_t^{(i)}, o_{<s}^{(i)}\right)}.
\label{eq:weight_ratios}
\end{equation}
This joint formulation compares direct and reflective rollouts within the same group. With binary rewards, group normalization changes only advantage magnitude: in every non-degenerate group, correct rollouts retain positive advantages and incorrect rollouts retain negative advantages regardless of interface, so Direct Reasoning is not systematically disadvantaged.

\begin{table*}[t]
\centering
\setlength{\tabcolsep}{2.0pt}
\renewcommand{\arraystretch}{1.0}
\resizebox{\textwidth}{!}{%
\begin{tabular}{lccccccc|cccc}
\toprule
\textbf{Model / Method} & \multicolumn{7}{c}{\textbf{In-Distribution Performance}} & \multicolumn{4}{c}{\textbf{Out-of-Distribution Performance}} \\
\cmidrule(lr){2-8} \cmidrule(lr){9-12}
 & \textbf{AIME24} & \textbf{AIME25} & \textbf{AMC} & \textbf{MATH-500} & \textbf{Minerva} & \textbf{Olympiad} & \textbf{Avg.} & \textbf{ARC-c} & \textbf{GPQA}$^{*}$ & \textbf{MMLU-Pro} & \textbf{Avg.} \\
 \midrule
Qwen2.5-Math-7B
  & 11.4 & 4.9 & 31.3 & 43.6 & 7.4 & 15.6 & 19.0 & 18.2 & 11.1 & 16.9 & 15.4 \\
Qwen2.5-Math-7B-Instruct
  & 12.9 & 10.2 & 48.5 & 80.4 &  32.7 & 41.0 & 37.6 & 70.3 & 24.7 & 34.1  & 43.0  \\
\midrule
\multicolumn{12}{c}{\textit{Previous Zero RLVR Methods}} \\
\midrule
PRIME-Zero [TMLR'25]
  & 17.0 & 12.8 & 54.0 & 81.4 &  39.0 & 40.3 & 40.7 & 73.3 & 18.2 & 32.7 & 41.4 \\
SimpleRL-Zero [CoLM'25]
  & 27.0 & 6.8 & 54.9 & 76.0 & 25.0 &  34.7 & 37.4 & 30.2 & 23.2 &  34.5 & 29.3 \\
OpenReasoner [NeurIPS'25]
  & 16.5 & 15.0 &  52.1 & 82.4 & 33.1 & 47.1  & 41.0 & 66.2  & 29.8  & 58.7 & 51.6 \\
  \midrule
\multicolumn{12}{c}{\textit{Supervised Learning Methods}} \\
\midrule
SFT
  & 17.2 & 15.1 &  49.8 & 72.1 & 31.2  & 33.2 & 36.4 & 35.1 & 16.9 & 29.5 & 27.2 \\
SFT-KL
  & 11.4 & 11.2 & 36.1 & 52.9 & 21.4 & 27.5 & 26.8 & 30.1 & 15.2 & 21.1 & 22.1 \\
\midrule
\midrule
\multicolumn{12}{c}{\textit{RLVR with ReflectRL}} \\
\midrule
GRPO [Nature'25]
  & 22.2 & 10.5 & 54.3 & 77.2 & 19.1 & 38.5 & 37.0 & 22.5 & 15.2 & 25.0 & 20.9 \\
\quad$\hookrightarrow$~+ EchoRL [ICML'26] & 23.4 & 11.1 & 55.2 & 77.1 & 24.6 & 39.8 & 38.5 & 33.8 & 15.8 & 27.9 & 25.8 \\
\quad$\hookrightarrow$~+ \textbf{ReflectRL} & \textbf{25.5} & \textbf{12.8} & \textbf{58.7} & \textbf{77.2} & \textbf{37.5} & \textbf{43.0} & \textbf{42.4}\textsuperscript{\scriptsize (+5.4)} & \textbf{65.4} & \textbf{17.7} & \textbf{36.7} & \textbf{40.0}\textsuperscript{\scriptsize (+19.1)} \\
\cmidrule(lr){1-12}
  DAPO [NeurIPS'25]
   & 22.9 & 11.8 & 56.5 & 76.4 & 22.4 & 40.2 & 38.4 & 30.8 & 16.2 & 28.8 & 25.3 \\
\quad$\hookrightarrow$~+ EchoRL [ICML'26] & 25.3 & 11.8 & 57.3 & 77.6 & 26.2 & 40.4 & 39.8 & 37.2 & 15.9 & 31.3 & 28.1 \\
\quad$\hookrightarrow$~+ \textbf{ReflectRL} & \textbf{31.0} & \textbf{12.1} & \textbf{59.8} & \textbf{80.6} & \textbf{36.8} & \textbf{41.3} & \textbf{43.5}\textsuperscript{\scriptsize (+5.1)} & \textbf{57.4} & 14.7 & \textbf{39.1} & \textbf{37.0}\textsuperscript{\scriptsize (+11.7)} \\
  \midrule
\multicolumn{12}{c}{\textit{On-Policy Distillation with ReflectRL (Student: Qwen2.5-3B-Instruct)}} \\
\midrule
OPD
  & 6.2 & 2.8 & 34.0 & 62.8 & 24.8 & 29.8 & 26.7 & 22.5 & 0.8 & 41.6 & 21.6 \\
\quad$\hookrightarrow$~+ \textbf{ReflectRL} & \textbf{8.1} & \textbf{4.5} & \textbf{36.5} & \textbf{64.6} & \textbf{27.9} & \textbf{30.5} & \textbf{29.0}\textsuperscript{\scriptsize (+2.3)} & \textbf{77.6} & \textbf{1.5} & 38.1 & \textbf{39.1}\textsuperscript{\scriptsize (+17.5)} \\
\bottomrule
\end{tabular}%
}
\caption{Overall in-distribution (ID) and out-of-distribution (OOD) performance comparison on Qwen2.5-Math-7B. ReflectRL consistently boosts existing on-policy post-training baselines (GRPO, DAPO, and OPD) in both ID and OOD scenarios, achieving competitive performance compared to previous zero-RL paradigms under a lightweight training recipe.}
\label{tab:main}
\end{table*}
\begin{figure*}[t!]
\centering
\includegraphics[width=0.32\textwidth]{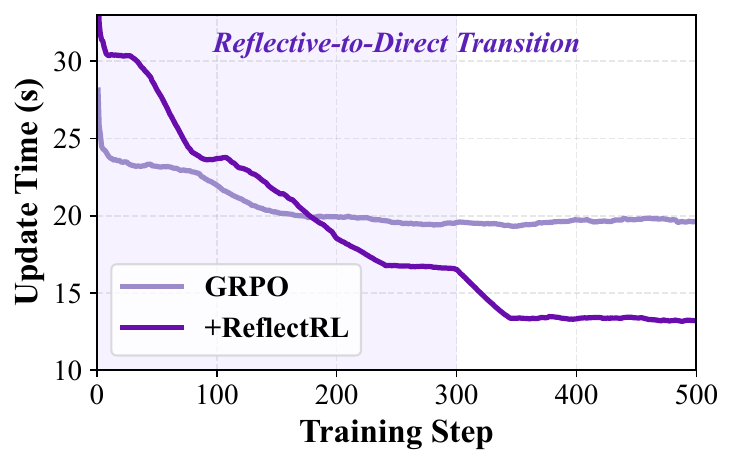}
\hfill
\includegraphics[width=0.32\textwidth]{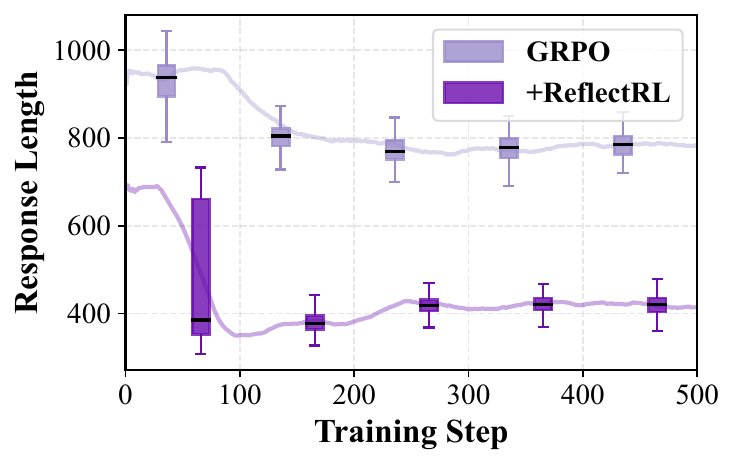}
\hfill
\includegraphics[width=0.32\textwidth]{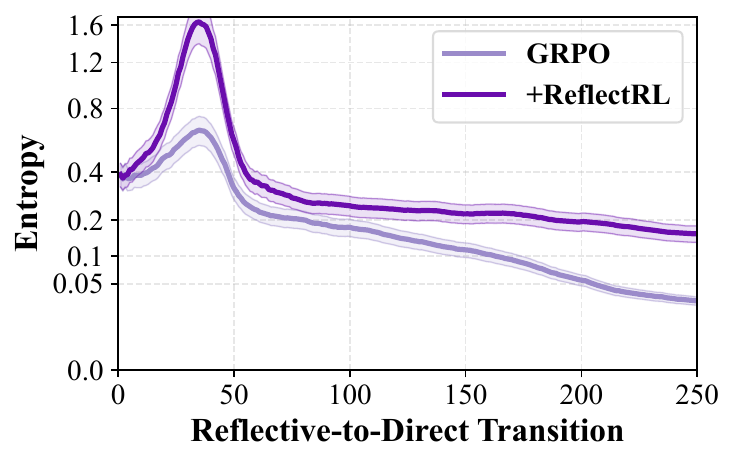}
\caption{Training dynamics comparison on Qwen2.5-Math-7B. (a) ReflectRL shifts computation from expensive search to cheap context processing, achieving higher update efficiency than GRPO. (b) ReflectRL achieves higher performance with more concise reasoning lengths. (c) Compared to GRPO, ReflectRL maintains active policy exploration and prevents premature entropy collapse.}
\label{fig:dynamics}
\end{figure*}
\subsection{ReflectRL in OPD}
\label{sec:opd_instantiation}
In On-Policy Distillation (OPD), ReflectRL utilizes the GNT as teacher-side privileged information to guide student training.
OPD generates one rollout per query, so we apply $g(t)$ across a batch of $B$ queries and set $K_t^{\mathrm{batch}}=\operatorname{round}(B g(t))$.
For sample $b$, the student generates $o^{(b)} \sim \pi_\theta(\cdot \mid x^D(q^{(b)}))$. The teacher context $\tilde{x}_t^{(b)}$ is reflective for $K_t^{\mathrm{batch}}$ samples and direct otherwise. Let
$p_{b,s}^{\mathrm{student}}=\pi_\theta(\cdot \mid x^D(q^{(b)}),o_{<s}^{(b)})$ and
$p_{b,s}^{\mathrm{teacher}}=\pi_{\mathrm{teacher}}(\cdot \mid \tilde{x}_t^{(b)},o_{<s}^{(b)})$. Then
\begin{equation}
\mathcal{L}_{\mathrm{OPD}}^{\mathrm{ReflectRL}}(\theta)
=
\mathbb{E}\!\left[\frac{1}{B}\sum_{b=1}^{B}\sum_{s=1}^{|o^{(b)}|}
\mathbb{D}_{\mathrm{KL}}\!\left(
p_{b,s}^{\mathrm{student}} \,\parallel\, p_{b,s}^{\mathrm{teacher}}
\right)
\right],
\label{eq:opd_reflectrl}
\end{equation}
where the reflective context is $x^R(q^{(b)},o^{-(b)})$ and the direct context is $x^D(q^{(b)})$.
The GNT $o^{-(b)}$ is provided strictly as teacher-side privileged context.
This formulation ensures that the student model never observes the GNT, retaining the direct reasoning interface $x^D(q)$ throughout training.
The teacher policy $\pi_{\mathrm{teacher}}$, guided by the GNT, provides a higher-quality target distribution $p_{b,s}^{\mathrm{teacher}}$ by actively reflecting on the error prefix.
By matching this distribution, the student distills the underlying error-avoidance and correction capabilities into its direct reasoning pathway, eliminating any GNT dependence during inference.

\subsection{Computational Overhead}
\label{sec:computational_overhead}
ReflectRL is highly computationally efficient because Golden Negative Trajectories are pre-generated offline, requiring no online expert queries. It retains the original rollout and verifier budgets, maintaining the same policy update rate as baseline training. The only extra overhead is prefilling GNT tokens in the reflective prompt, which accounts for a negligible fraction of training time and decreases to zero as $g(t)$ decays.

\begin{figure*}[t]
\centering
\includegraphics[width=\textwidth]{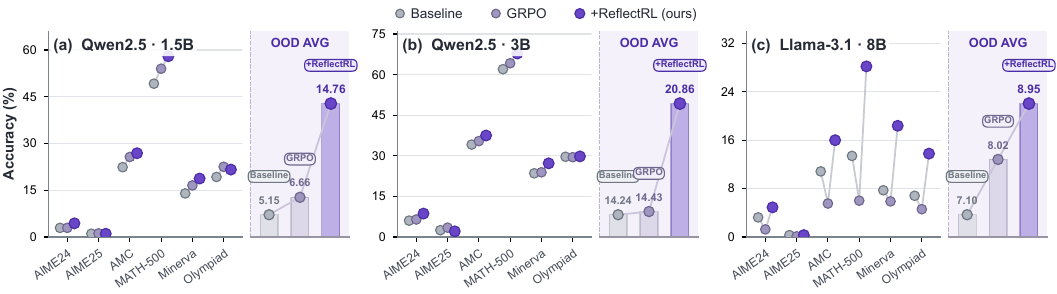}
\caption{Performance comparison of Baseline, GRPO, and ReflectRL (+ReflectRL) across Llama-3.1-8B-Instruct, Qwen2.5-1.5B-Instruct, and Qwen2.5-3B-Instruct on 9 ID and OOD benchmarks. ReflectRL consistently outperforms standard GRPO across various model sizes and families, showing strong generalization capabilities.}
\label{fig:performance_comparison}
\end{figure*}

\section{Experiments}\label{sec:exp}

In this section, we conduct extensive experiments to address the following research questions: (\textbf{RQ1: Overall Effectiveness}) How does ReflectRL perform when integrated into existing on-policy post-training methods? (\textbf{RQ2: Computational Efficiency}) Does ReflectRL introduce significant computational overhead? (\textbf{RQ3: Mechanistic Validity}) How sensitive is ReflectRL to the proposed reflective mechanism and the choice of policy transition schedule?

\subsection{Experiment Setup}\label{sec:exp-setup}

\paragraph{Datasets and Benchmarks} We construct OpenR1-GNT-69k from a subset of OpenR1-Math-220k. The prompts are drawn from NuminaMath 1.5, and the candidate trajectories are generated by DeepSeek-R1. We retain trajectories that are verified as incorrect by Math-Verify, yielding 69k Golden Negative Trajectories for training.
To evaluate both in-domain learning and transfer, we use 9 widely used reasoning benchmarks. Six benchmarks are in-distribution mathematical reasoning tasks: AIME 2024/2025, AMC, MATH-500 \cite{hendrycks2021measuringmathematicalproblemsolving}, Minerva \cite{lewkowycz2022solvingquantitativereasoningproblems}, and OlympiadBench \cite{he2024olympiadbenchchallengingbenchmarkpromoting}. Three benchmarks evaluate out-of-distribution generalization: ARC-c \cite{clark2018thinksolvedquestionanswering}, GPQA-Diamond \cite{rein2023gpqagraduatelevelgoogleproofqa}, and MMLU-Pro \cite{wang2024mmluprorobustchallengingmultitask}. For AIME 2024, AIME 2025, and AMC, we report avg@32 because the test sets are small; for the remaining benchmarks, we report pass@1.

\paragraph{Baselines}
We evaluate ReflectRL by integrating it into four representative on-policy training methods: GRPO \cite{DeepSeekAI2025DeepSeekR1IR}, DAPO \cite{yu2026dapo}, EchoRL \cite{bi2026echorl}, and OPD \cite{lu2025onpolicydistillation}. EchoRL recovers learning signals from advantage-degenerated rollout groups by extracting entropy-selected clips from verified-success rollouts and feeding them back as auxiliary supervision. We also report results from prior work, including: (1) SFT on an equivalent amount of verified correct trajectories extracted from OpenR1 for fair comparison; (2) SFT with a KL-divergence constraint incorporated into the objective (SFT-KL); and three RL baselines: (3) SimpleRL-Zero \cite{zeng2025simplerlzoo}, which applies GRPO to approximately 24k mathematical samples from GSM8K and MATH; (4) OpenReasoner-Zero \cite{hu2025openreasonerzero}, a PPO-based method trained on 129k multi-source samples, including AIME; and (5) PRIME-Zero \cite{cui2025processreinforcementimplicitrewards}, which performs policy rollouts on 150k NuminaMath queries using implicit process rewards and outcome labels.

\paragraph{Models} We consider 4 representative LLMs ranging from 1.5B to 8B parameters from the Qwen2.5 \cite{qwen2025qwen25technicalreport} and LLaMA-3.1 \cite{grattafiori2024llama3herdmodels} families, and train them on OpenR1-GNT-69k. For OPD, we use Qwen2.5-3B-Instruct as the student model and Qwen2.5-7B-Instruct as the expert model.

\paragraph{Parameter Configurations} We implement ReflectRL in verl\footnote{\url{https://github.com/volcengine/verl}}. We use a rollout batch size of 128 and an update batch size of 64. During rollout generation, we sample 8 responses for each on-policy question. We shuffle multiple-choice options to reduce contamination risk. For evaluation, we set the temperature to 0.6. For OPD, we use reverse KL and a rollout batch size of 1024. All experiments are run on 16 H200 GPUs. Additional training and evaluation details are provided in Appendix~E.

\begin{figure}[!t]
\centering
\includegraphics[width=\linewidth]{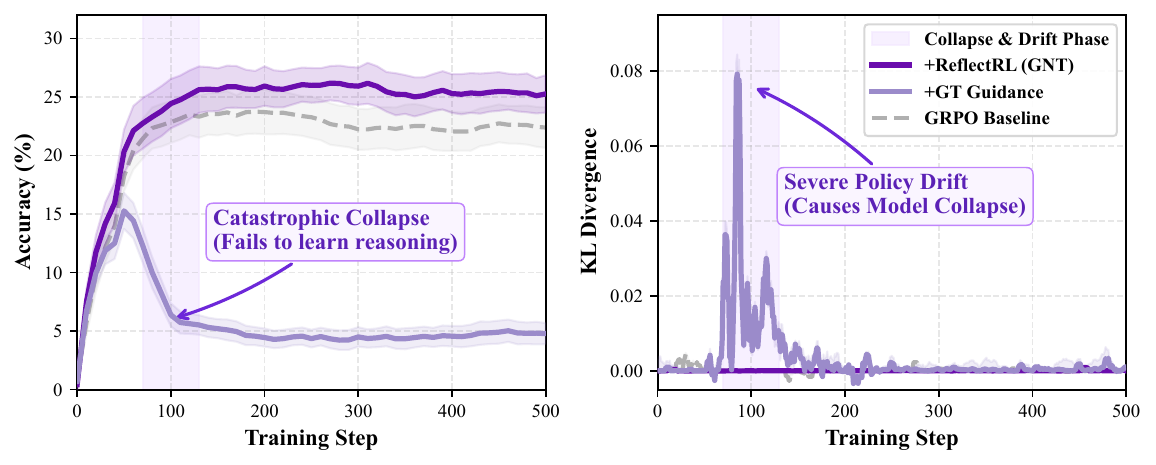}
\caption{Ablation analysis on Qwen2.5-Math-7B. (a) Accuracy dynamics under different guidance source. (b) Policy drift measured by KL divergence. Highlighted region (steps 70--130) demonstrates the clear causal link between severe policy drift and catastrophic collapse under +GT Guidance, whereas ReflectRL (GNT) stabilizes policy updates.}
\label{fig:ablation}
\end{figure}
\subsection{RQ1: Overall Effectiveness}
We integrate ReflectRL into various on-policy baselines to evaluate its impact on reasoning performance and generalization across different model scales and domains.

\textbf{Takeaway 1: ReflectRL delivers consistent improvements over various baselines across diverse architectures and domains.}
Table~\ref{tab:main} and Figure~\ref{fig:performance_comparison} show that ReflectRL consistently improves GRPO, DAPO, and OPD across Qwen and LLaMA models from 1.5B to 8B parameters. OpenR1-GNT-69k and ARC-c have no data overlap \cite{huang2026think,limozin2026sftthenrl}. The substantial ARC-c gain therefore reflects transfer beyond the mathematical training distribution: its causal and logical consistency checks and distractor rejection align with the reflect--repair--solve process induced by GNTs, indicating that ReflectRL acquires domain-general reasoning capabilities.

\textbf{Takeaway 2: ReflectRL elicits more concise reasoning through Reflective-to-Direct Policy Transition.}
Reasoning length can serve as a behavioral indicator of reasoning efficiency, where shorter successful trajectories often reflect more direct and selective problem solving \cite{Wu2025WhenMI,hassid2026dont}. As shown in Figure~\ref{fig:dynamics}(b), GRPO produces responses exceeding 800 tokens by step 500, whereas ReflectRL remains near 420 tokens while achieving higher accuracy. This suggests that Reflective-to-Direct Policy Transition effectively elicits a more concise reasoning behavior, allowing the model to reach correct solutions with substantially fewer tokens.

\textbf{Takeaway 3: ReflectRL effectively mitigates entropy collapse and sustains continuous learning.}
Policy entropy provides a useful view of whether training remains exploratory or collapses to a narrow response mode. In Figure~\ref{fig:dynamics}(c), GRPO entropy falls from 0.97 to below 0.03 by step 250. ReflectRL remains around 0.15 at the same step, roughly five times higher than GRPO. This sustained entropy correlates strongly with continuous performance growth, indicating that ReflectRL effectively internalizes this reflective reasoning capability into the model parameters, enabling sustained exploration throughout training.

\begin{figure}[!t]
\centering
\includegraphics[width=\linewidth]{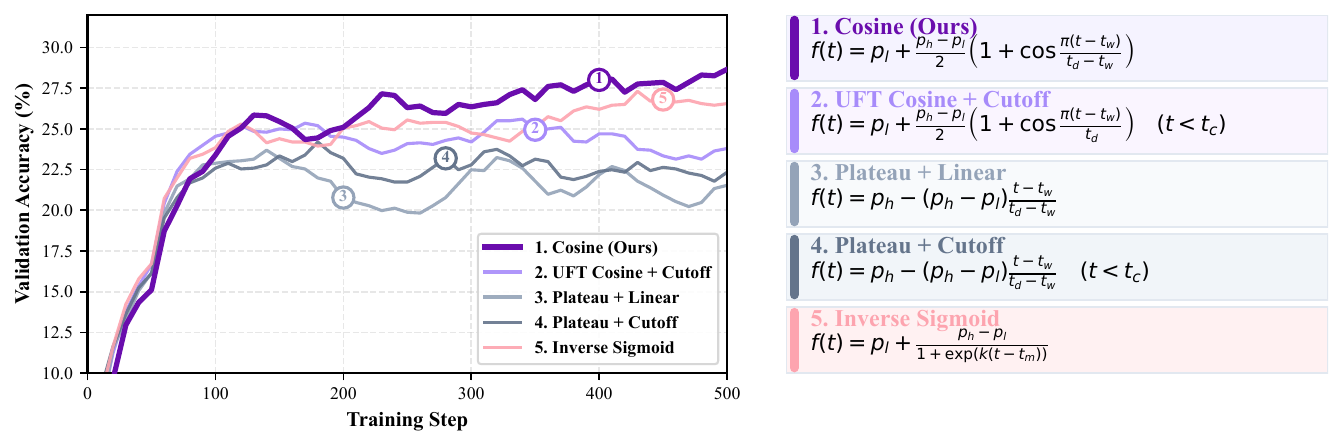}
\caption{Comparison of different policy transition kernels on Qwen2.5-Math-7B. (Left) Validation accuracy over training steps with smoothed curves. (Right) Corresponding generalized mathematical equations color-coded to match the curves (specific parameters are detailed in Appendix~C).}
\label{fig:transition_comparison}
\end{figure}
\subsection{RQ2: Computational Efficiency}

To evaluate the efficiency of ReflectRL, we analyze its training dynamics and computational profile in Figure~\ref{fig:dynamics}.

\textbf{Takeaway 4: ReflectRL reduces training time through shorter rollout generation.}
Figure~\ref{fig:dynamics}(a) shows that the update time of ReflectRL decreases throughout the Reflective-to-Direct Policy Transition, eventually stabilizing at approximately 13 seconds per step, compared with around 20 seconds for GRPO. Although reflective rollouts introduce additional GNT tokens during prefilling, this cost is offset by the substantially shorter continuations shown in Figure~\ref{fig:dynamics}(b), which reduce the more expensive autoregressive decoding workload. Since the GNTs are generated offline and the rollout and verifier budgets remain unchanged, ReflectRL ultimately achieves higher reasoning performance with lower per-step training cost.
ReflectRL introduces no auxiliary model, trainable parameters, gradients, or optimizer states, and therefore adds no persistent memory overhead. Its only additional cost is the temporary KV cache for reflective context; as training proceeds, the savings from substantially shorter rollout generation increasingly outweigh this cost, while the Reflective-to-Direct Policy Transition gradually reduces the extra KV-cache usage to zero.

\subsection{RQ3: Mechanistic Validity}\label{sec:exp-framework}

\paragraph{Source of Reflection}
We progressively isolate the source of effective reflection from three perspectives: trajectory provenance, trajectory components, and training stability. First, Figure~\ref{fig:reflection_advantage}(a) shows that not all failures are equally useful. GNTs produce a large and steadily increasing Reflection Gain, whereas self-generated failures provide only limited benefits, and weak-model failures or length-matched mismatched GNTs yield negative gains. This rules out the possibility that reflection simply benefits from additional context or arbitrary incorrect reasoning; the failure must be both high-quality and aligned with the target problem. Second, Figure~\ref{fig:reflection_advantage}(c) further localizes the useful signal within a GNT. The valid prefix provides a strong reasoning scaffold, while the localized error region supplies a concrete target for correction; combining both yields the largest gain. In contrast, shuffling the trajectory or retaining only the final answer reduces performance below Direct Reasoning, showing that coherent reasoning structure is essential. Finally, Figure~\ref{fig:ablation} shows that correct expert trajectories(GT) are not necessarily better guidance. Although GT guidance brings an early improvement, it induces severe policy drift and eventually collapses, whereas GNT guidance remains stable throughout training. Together, these results identify GNTs as a particularly effective source of reflection: they are more informative than low-quality failures, yet still require the policy to diagnose and repair reasoning rather than directly imitate an expert solution.

\paragraph{Influence of Policy Transition Kernels}
Figure~\ref{fig:transition_comparison} shows that smooth transition kernels consistently yield stronger and more stable performance than abrupt or overly rapid schedules. The cosine kernel achieves the best final accuracy, while the inverse-sigmoid variant follows a similar trend, suggesting that the exact functional form is less important than maintaining a gradual shift from Reflective Reasoning to Direct Reasoning. In contrast, sharper transitions weaken performance by removing reflective guidance before the policy has fully internalized the learned behavior.

\section{Conclusion}

We introduce ReflectRL, a lightweight framework that learns from GNTs through reflection rather than imitation. Reflective-to-Direct Policy Transition transfers this behavior to direct reasoning while preserving the original on-policy training pipeline. Experiments across RLVR and OPD show consistent gains in accuracy, efficiency, and training stability.

\clearpage
\bibliographystyle{unsrtnat}
\bibliography{aaai2027}

\clearpage
\beginappendix

\section{Prompt Templates}\label{app:templates}
Here we present the prompt templates used in ReflectRL. During on-policy training, all model inputs are structured using a shared system prompt that guides the policy to generate systematic step-by-step reasoning within a designated thought container.

\begin{figure*}[!htbp]
    \centering
    \includegraphics[width=0.9\linewidth]{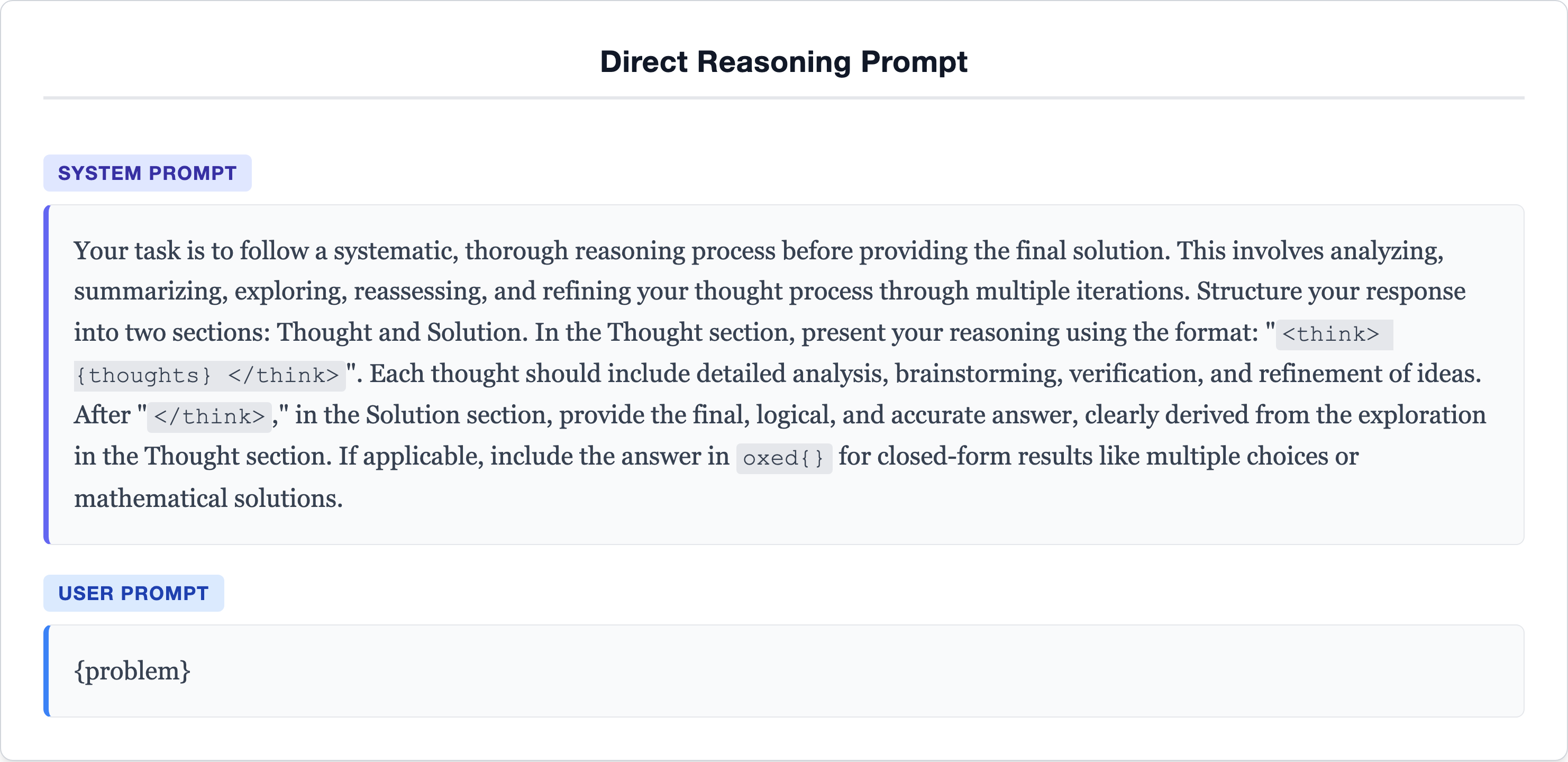}
    \caption{The format of the Direct Reasoning Prompt ($\text{Chat\_Temp}_D$).}
    \label{fig:prompt_direct}
\end{figure*}

\paragraph{Direct Reasoning}
The direct reasoning interface applies $\text{Chat\_Temp}_D$. We design a structured template to present the original problem alongside the system prompt, as shown in Figure~\ref{fig:prompt_direct}.

\begin{figure*}[!htbp]
    \centering
    \includegraphics[width=0.9\linewidth]{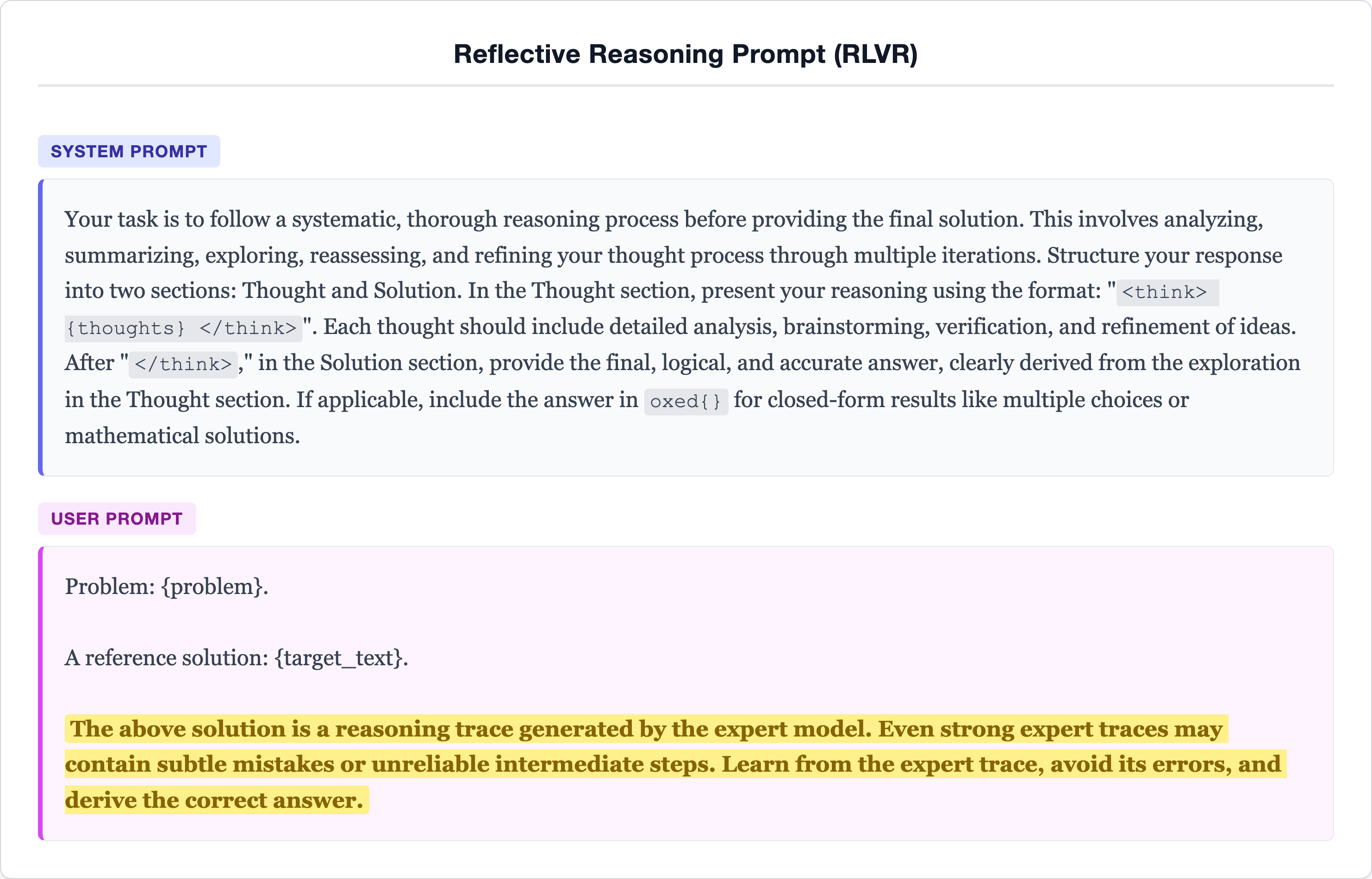}
    \caption{The format of the Reflective Reasoning Prompt ($\text{Chat\_Temp}_R$).}
    \label{fig:prompt_reflective}
\end{figure*}

\paragraph{Reflective Reasoning}
The reflective reasoning interface applies $\text{Chat\_Temp}_R$. This template presents a pre-generated Golden Negative Trajectory (GNT) as contextual information, prompting the policy to identify its errors and repair the reasoning process, as shown in Figure~\ref{fig:prompt_reflective}.


\begin{figure*}[t]
    \centering
    \includegraphics[width=0.9\linewidth]{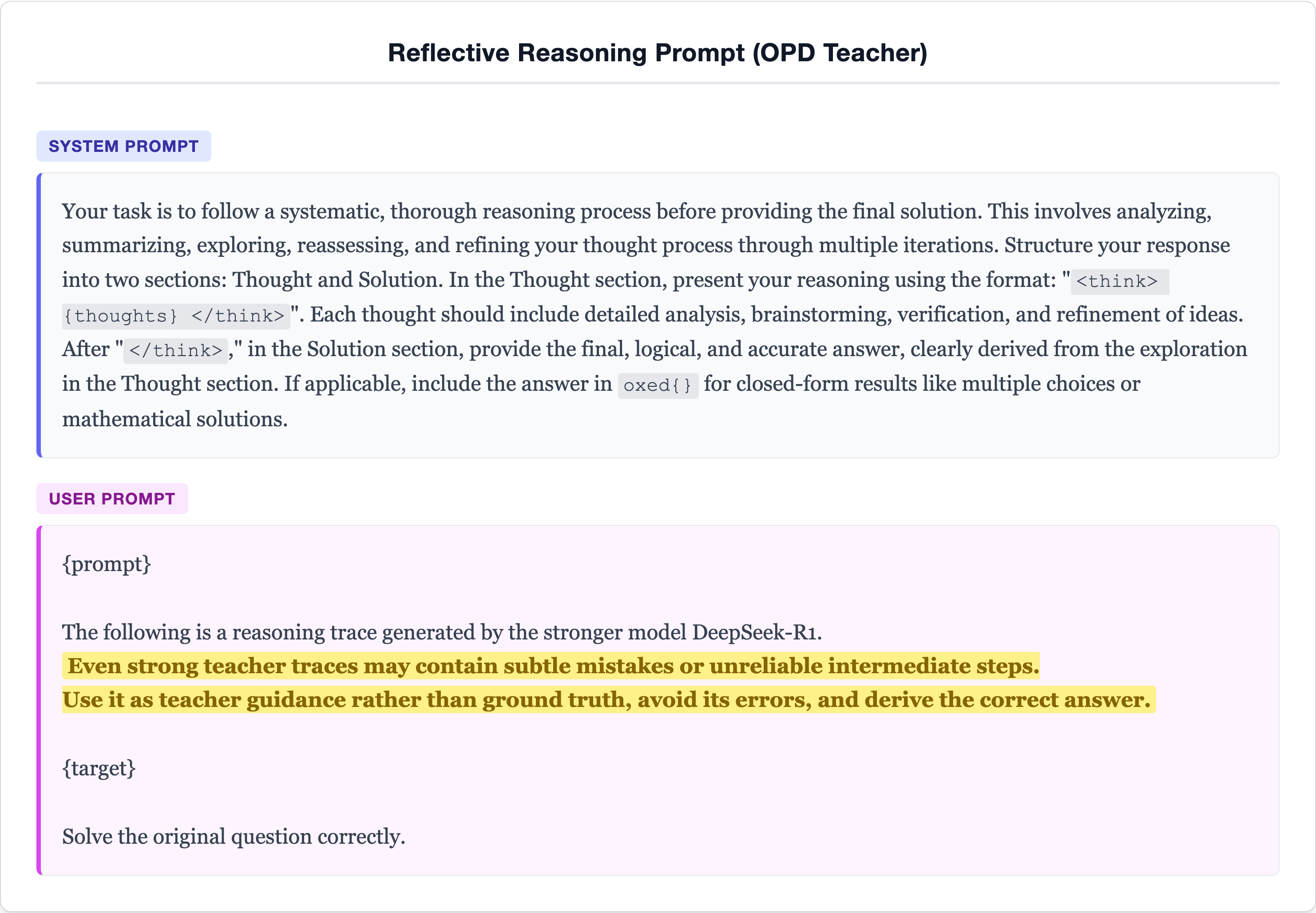}
    \caption{The format of the Reflective Reasoning Prompt for the OPD Teacher.}
    \label{fig:prompt_opd}
\end{figure*}

\paragraph{Reflective Reasoning in OPD}
In On-Policy Distillation (OPD), the expert teacher model is guided by a specific reflective prompt to process the GNT as privileged context. As shown in Figure~\ref{fig:prompt_opd}, this template instructs the teacher to view the trace as guidance rather than ground truth, avoid its errors, and derive the correct solution for the student to match.

\section{Detailed Evaluation Results}\label{app:evaluation_results}
We report the exact evaluation accuracy numbers corresponding to Figure~4 in the main paper.

\begin{table*}[!htbp]
\centering
\setlength{\tabcolsep}{2.0pt}
\renewcommand{\arraystretch}{1.0}
\resizebox{\textwidth}{!}{%
\begin{tabular}{lccccccc|cccc}
\toprule
\textbf{Model / Method} & \multicolumn{7}{c}{\textbf{In-Distribution Performance}} & \multicolumn{4}{c}{\textbf{Out-of-Distribution Performance}} \\
\cmidrule(lr){2-8} \cmidrule(lr){9-12}
 & \textbf{AIME24} & \textbf{AIME25} & \textbf{AMC} & \textbf{MATH-500} & \textbf{Minerva} & \textbf{Olympiad} & \textbf{Avg.} & \textbf{ARC-c} & \textbf{GPQA}$^{*}$ & \textbf{MMLU-Pro} & \textbf{Avg.} \\
 \midrule
\textit{Qwen2.5-1.5B-Instruct} & & & & & & & & & & & \\
Baseline & 2.92 & 1.04 & 22.40 & 49.20 & 13.97 & 19.26 & 18.13 & 2.90 & 1.01 & 11.53 & 5.15 \\
GRPO & 2.92 & 1.15 & 25.68 & 54.00 & 16.54 & 22.52 & 20.47 & 0.34 & 1.01 & 18.63 & 6.66 \\
+ReflectRL & 4.37 & 1.04 & 26.88 & 58.00 & 18.75 & 21.63 & 21.78 & 18.60 & 2.53 & 23.15 & 14.76 \\
\midrule
\textit{Qwen2.5-3B-Instruct} & & & & & & & & & & & \\
Baseline & 6.04 & 2.50 & 34.15 & 62.00 & 23.53 & 29.63 & 26.31 & 0.43 & 0.51 & 41.77 & 14.24 \\
GRPO & 6.46 & 3.44 & 35.43 & 64.20 & 23.90 & 29.48 & 27.15 & 1.71 & 1.01 & 40.58 & 14.43 \\
+ReflectRL & 8.65 & 2.08 & 37.50 & 67.80 & 27.21 & 29.78 & 28.84 & 19.37 & 1.01 & 42.20 & 20.86 \\
\midrule
\textit{Llama-3.1-8B-Instruct} & & & & & & & & & & & \\
Baseline & 3.23 & 0.31 & 10.84 & 13.40 & 7.72 & 6.81 & 7.05 & 0.00 & 0.00 & 21.30 & 7.10 \\
GRPO & 1.25 & 0.10 & 5.53 & 6.00 & 5.88 & 4.59 & 3.89 & 0.00 & 0.00 & 24.05 & 8.02 \\
+ReflectRL & 4.90 & 0.31 & 16.00 & 28.20 & 18.38 & 13.78 & 13.60 & 0.17 & 0.00 & 26.67 & 8.95 \\
\bottomrule
\end{tabular}%
}
\caption{Detailed evaluation results of Baseline, GRPO, and ReflectRL (+ReflectRL) across Qwen2.5-1.5B-Instruct, Qwen2.5-3B-Instruct, and Llama-3.1-8B-Instruct on 9 in-distribution (ID) and out-of-distribution (OOD) benchmarks.}
\label{tab:detailed_performance_comparison}
\end{table*}

\begin{figure*}[t]
    \centering
    \includegraphics[width=0.9\linewidth]{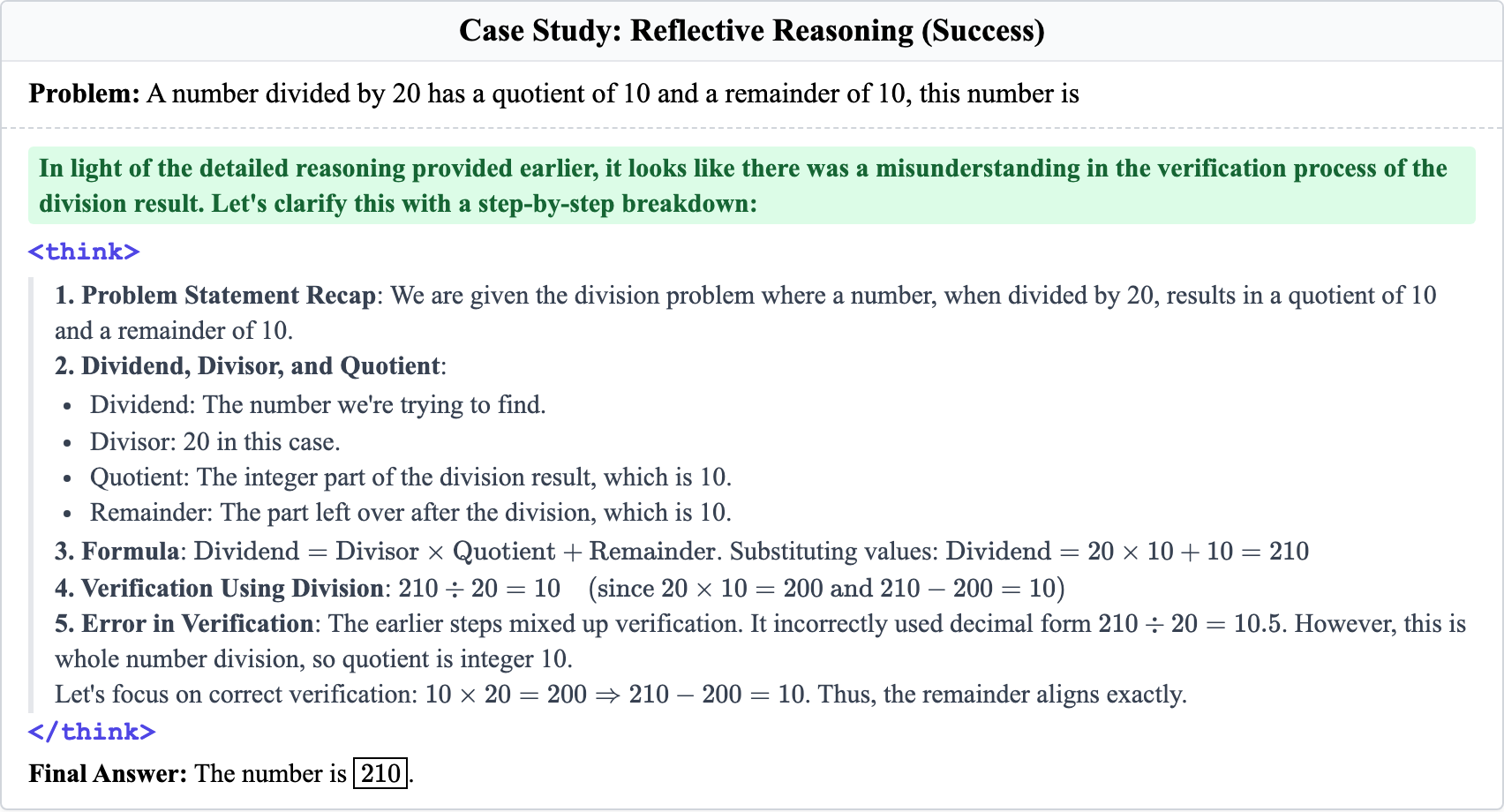}
    \caption{Case Study: Reflective Reasoning (Success).}
    \label{fig:case_study_reflect}
\end{figure*}

\begin{figure*}[t]
    \centering
    \includegraphics[width=0.9\linewidth]{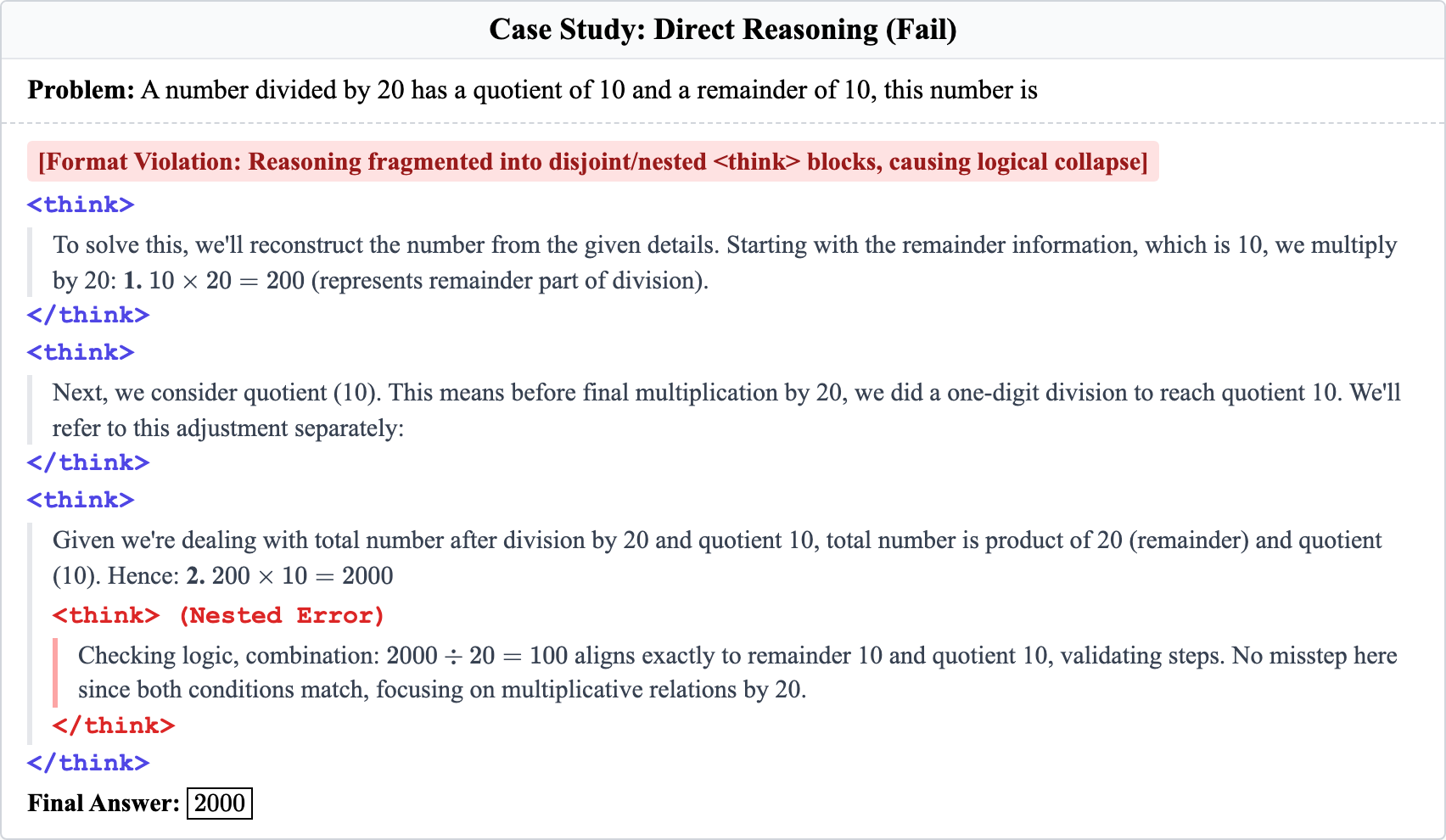}
    \caption{Case Study: Direct Reasoning (Fail).}
    \label{fig:case_study_direct}
\end{figure*}
\section{Policy Transition Schedulers}\label{app:policy_transition_kernels}
To transition the guidance probability $g(t)$ from high reflection to standard RL over the training steps $t$, we define a set of schedulers. The generalized mathematical expressions for the six evaluated transition kernels are formulated below, governed by hyper-parameters: initial high probability $p_h$, final low probability $p_l$, warm-up steps $t_w$, decay horizon $t_d$, and a hard cutoff step $t_c$. In our experiments, we set $p_h = 0.5$, $p_l = 0.05$ (or $0.0$ for cutoff variants), $t_w = 50$, $t_d = 250$, and $t_c = 300$. To simplify the notation, we denote the probability drop as $\Delta p = p_h - p_l$.

\paragraph{1. UFT Cosine + Cutoff}
\begin{equation*}
g(t)=
\begin{cases}
p_h, & t < t_w, \\[4pt]
p_l + \frac{\Delta p}{2} \left[1+\cos\left(\pi \frac{t-t_w}{t_d}\right)\right], & t_w \leq t < t_c, \\[4pt]
0, & t \geq t_c.
\end{cases}
\end{equation*}

\paragraph{2. Linear}
\begin{equation*}
g(t)=
\begin{cases}
p_h - \Delta p \cdot \frac{t}{t_d}, & t < t_d, \\[4pt]
p_l, & t \geq t_d.
\end{cases}
\end{equation*}

\paragraph{3. Cosine}
\begin{equation*}
g(t)=
\begin{cases}
p_h, & t < t_w, \\[4pt]
p_l + \frac{\Delta p}{2} \left[1+\cos\left(\pi \frac{t-t_w}{t_d - t_w}\right)\right], & t_w \leq t < t_d, \\[4pt]
p_l, & t \geq t_d.
\end{cases}
\end{equation*}

\paragraph{4. Plateau + Linear}
\begin{equation*}
g(t)=
\begin{cases}
p_h, & t < t_w, \\[4pt]
p_h - \Delta p \cdot \frac{t-t_w}{t_d - t_w}, & t_w \leq t < t_d, \\[4pt]
p_l, & t \geq t_d.
\end{cases}
\end{equation*}

\paragraph{5. Plateau + Cutoff}
\begin{equation*}
g(t)=
\begin{cases}
p_h, & t < t_w, \\[4pt]
p_h - \Delta p \cdot \frac{t-t_w}{t_d - t_w}, & t_w \leq t < t_d, \\[4pt]
p_l, & t_d \leq t < t_c, \\[4pt]
0, & t \geq t_c.
\end{cases}
\end{equation*}

\paragraph{6. Inverse Sigmoid}
\begin{equation*}
g(t)=p_l+\frac{\Delta p}{1+\exp\left(k(t-t_m)\right)}
\end{equation*}
where $t_m = \frac{t_w+t_d}{2}$ and $k$ controls the steepness of the curve.

\section{Related Work}\label{app:related_work}

\paragraph{Reinforcement Learning with Verifiable Rewards (RLVR)}
Recently, large language models and multi-modal LLMs (MLLMs) have demonstrated remarkable capabilities across various complex reasoning tasks \cite{bi-etal-2025-llava,Bi2025PRISMSI,zhao2026nl2codestructuredsurveymultimodal,zhang2023spot,Zhang2025TheLO,yang2026drdocbenchcomprehensivebenchmark,yang2026alignsaeconceptalignedsparseautoencoders,jiang2025koreenhancingknowledgeinjection,Wang_Bi_Pirk_Ma_2026,wan2025magicwordssharpnessawareprompt,wan2025hyperion,rong2026backdoor,peng2025visualinputcompressedvisual,jiang2025minedprobingupdatingmultimodal}. To further enhance these capabilities, Reinforcement Learning with Verifiable Rewards has emerged as a predominant paradigm for post-training on reasoning tasks \cite{lightman2024let,DeepSeekAI2025DeepSeekR1IR}. Unlike traditional RLHF which relies on reward models trained on human preferences, RLVR leverages rule-based verifiers or deterministic outcome signals (e.g., matching a mathematical answer or passing unit tests) to provide unambiguous reward signals \cite{zelikman2022star, wang2024mathcoder}. Recent advancements in this space, such as Group Relative Policy Optimization (GRPO), eliminate the need for an external critic model by computing advantages relative to a sampled group of rollouts \cite{DeepSeekAI2025DeepSeekR1IR}. While RLVR is highly effective at exploring and reinforcing correct reasoning paths, it heavily relies on the model's intrinsic capability to occasionally sample the correct answer. On highly complex problems where the model fails to generate any correct trajectory, the reward signal becomes sparse, leading to inefficient exploration and degraded policy updates \cite{bi2026echorl}. ReflectRL builds upon this paradigm by injecting Reflective Reasoning into the RLVR process, ensuring that even when direct exploration fails, the model can still receive dense and constructive signals by reflecting on Golden Negative Trajectories.

\paragraph{On-Policy Distillation (OPD)}
On-Policy Distillation aims to transfer knowledge from a stronger expert model (teacher) to a target model (student) using the student's own generated trajectories \cite{gu2024minillm,zhao2026selfdistilledreasoneronpolicyselfdistillation,lu2025onpolicydistillation}. By computing the Kullback-Leibler (KL) divergence between the teacher and student distributions on on-policy states, OPD mitigates the distribution shift and exposure bias issues typically encountered in offline Supervised Fine-Tuning (SFT) \cite{Zhang2025TheLO}. In standard OPD, the teacher model evaluates the student's reasoning path and provides a token-level distribution to guide the student. However, when the problem is challenging, the teacher might also lack sufficient context to provide optimal step-by-step guidance. ReflectRL extends the OPD framework by providing the Golden Negative Trajectory as privileged contextual information to the teacher model. This allows the teacher to perform Reflective Reasoning and generate higher-quality target distributions, while the student model distills this error-correction capability directly into its Direct Reasoning pathway without relying on the external trajectory during inference.

\paragraph{Learning from Expert Trajectories}
The use of expert trajectories is a common strategy to improve reasoning capabilities in language models \cite{liu2026uft,NEURIPS2025_a9d5c33e}. Traditional methods primarily treat successful expert trajectories as positive demonstrations to be imitated via behavioral cloning \cite{liu2026uft}. In the context of on-policy training, recent works have explored utilizing correct expert trajectories to initialize policy search or to serve as high-reward anchors to guide rollout generation \cite{NEURIPS2025_a9d5c33e}. However, these approaches uniformly discard expert failures as unusable negative samples. Some studies have investigated learning from negative examples or self-generated mistakes \cite{welleck2022generatingsequenceslearningselfcorrect}, but these typically focus on simple contrastive objectives or filtering heuristics. In contrast, ReflectRL uniquely identifies the \emph{Reflection Advantage} inherent in structured expert failures (Golden Negative Trajectories). Rather than discarding these failures or treating them as negative contrastive targets, ReflectRL leverages them as a contextual scaffold to elicit explicit Reflective Reasoning, demonstrating that reflecting on high-quality mistakes is often more effective than learning from scratch.

\section{Training Details}\label{app:training_details}
We implemented our training pipeline based on the open-source \texttt{verl} framework. Below are the key hyperparameters and configuration details used for ReflectRL in the on-policy rollout settings:
\begin{itemize}
    \item \textbf{Data Configuration}: We utilize a subset of the OpenR1-Math-220k dataset for training. The global training batch size is set to 128, and the validation batch size is 512. To ensure proper context parsing, we limit both the maximum prompt length and the maximum response length to 2048 tokens.
    \item \textbf{Rollout Generation}: For each training query, we sample $N=8$ responses using vLLM for high-throughput generation with a tensor parallel size of 2. The sampling temperature is set to 1.0 during training to encourage exploration and 0.6 during validation for stable evaluation.
    \item \textbf{PPO \& Optimization}: We employ the GRPO advantage estimator. The actor learning rate is set to $1 \times 10^{-6}$. For policy optimization, we utilize a PPO micro-batch size and mini-batch size of 64. To maintain continuous exploration, the entropy coefficient is set to 0.001. The KL penalty coefficient is set to 0.0 to prevent interference with the Reflective-to-Direct transition mechanism and allow the policy to freely shift away from the initial reference model.
    \item \textbf{Hardware and Distribution}: The models are trained across NVIDIA GPUs with DeepSpeed integration. Gradient checkpointing is enabled to save memory, and the Ulysses sequence parallel size is set to 1.
\end{itemize}

\section{Case Study}\label{app:case_study}
To further demonstrate the effectiveness of Reflective Reasoning, we present a detailed case study illustrating the behavioral differences between standard exploration and our proposed framework.

As shown in Figure~\ref{fig:case_study_direct}, the Direct Reasoning trajectory completely fails to solve a relatively straightforward arithmetic problem from scratch. Not only does the policy struggle to organize the steps cohesively, but it critically violates the structural constraint by fragmenting its reasoning into multiple disjoint and nested \texttt{<think>} blocks, leading to a logical collapse. Under standard RLVR, such a trajectory receives a zero reward, providing the model with an entirely uninformative gradient update that fails to guide it toward the correct format or logic. In sparse-reward environments, repeated failures of this nature lead to early convergence to sub-optimal policies.

In contrast, Figure~\ref{fig:case_study_reflect} demonstrates the power of the Reflection Advantage. When presented with the Golden Negative Trajectory as a contextual anchor, the policy successfully enters a Reflective Reasoning state. It explicitly acknowledges the shortcomings in the previous attempt (``The reference solution suggests there were some errors in the previous attempts. Recomputing:''), strictly adheres to the \texttt{<think>} reasoning structure, and correctly re-evaluates the arithmetic operations step-by-step. By learning to critique and reconstruct the solution from a high-quality mistake, ReflectRL converts a dead-end zero-reward exploration path into a dense, high-reward learning signal.

\end{document}